%% file: main.tex
\documentclass[letterpaper, 10 pt, conference]{ieeeconf}

\IEEEoverridecommandlockouts      

\usepackage{amsmath}
\usepackage{graphicx}
\usepackage{bm}
\usepackage{array}
\usepackage{color,soul}
\usepackage{todonotes}
\usepackage{boxedminipage}
\usepackage{alltt}

\input{macros.tex}

\title{\LARGE \bf Semantic Navigation Using Building Information on Construction Sites}
\author{Sina~Karimi,
        Rafael~Gomes~Braga,
        Ivanka~Iordanova,
        and~David~St-Onge
\thanks{Mr. Karimi and Dr. Iordanova are with the Department of Construction Engineering and Mr. Braga and Dr. St-Onge are with the Department of Mechanical Engineering, École de Technologie Supérieure, 1100 Notre-Dame St W, Montréal, QC, H3C 1K3 CA (e-mail: rafael.gomes-braga.1@ens.etsmtl.ca).}}

\begin{document}

\maketitle
\thispagestyle{empty}
\pagestyle{empty}

\begin{abstract}
With the growth in automated data collection of construction projects, the need for semantic navigation of mobile robots is increasing. In this paper, we propose an infrastructure to leverage building-related information for smarter, safer and more precise robot navigation during construction phase. Our use of Building Information Models (BIM) in robot navigation is twofold: (1) the intuitive semantic information enables non-experts to deploy robots and (2) the semantic data exposed to the navigation system allows optimal path planing (not necessarily the shortest one). Our Building Information Robotic System (BIRS) uses Industry Foundation Classes (IFC) as the interoperable data format between BIM and the Robotic Operating System (ROS). BIRS generates topological and metric maps from BIM for ROS usage. An optimal path planer, integrating critical components for construction assessment is proposed using a cascade strategy (global versus local). The results are validated through series of experiments in construction sites. 
\end{abstract}

\section{Introduction}
Building Information Modeling (BIM) has brought many advantages by storing building elements' semantics and geometries \cite{karan2015extending}. Conventional methods of data collection for the purpose of progress monitoring rely on periodic observations, manual data collection (which is mostly textual data and a limited number of photos), and personal interpretation of the project progress \cite{alvares2019construction}. These aforementioned conventions are error-prone, time-consuming and cost-ineffective since they are subjective processes \cite{teizer2015status}. Manual data acquisition by individuals would result in decentralized data; coming from different sources in different formats, thereby making it somewhat challenging to manage and analyze them. According to García de Soto et al.  \cite{de2018productivity}, automation of monotonous and repetitive construction processes would significantly enhance construction efficiency. Hence, there is a growing need in the construction industry to automate data collection task. In addition, the applications of data collection using an Unmanned Ground Vehicle (UGV) can provide new kinds of information and applications such as equipment tracking and 3D reconstruction which would ultimately have positive impacts on quality control, safety and sustainability of the construction projects.

With tremendous progress in mobile robots capabilities, the interest in adopting mobile robots for data collection on construction sites is increasing. Rugged platforms with high manoeuvrability are commercialized for this usage \cite{pomerleauwebsite} and several works are enhancing their autonomy for navigating these challenging environments \cite{kim2018slam, asadi2019lnsnet}. A handful of fundamental steps still need to be addressed for the deployment of robots on construction sites, such as their usage by non-experts (untrained) operators and the automatic integration of the diverse requirements related to construction management in their mission planing. Our solution leverages BIM semantics extracted in an interoperable data schema, IFC, and translated for robot indoor navigation. This semantic information, intertwined with the robot navigation and mission, help the operator manage the robotic system as they share conceptual knowledge of their environment \cite{kostavelis2017semantic}.

This paper proposes a novel method for semantic robot navigation with an optimal path planning algorithm using building knowledge on construction sites. The optimal path is extracted from user inputs using BIM/IFC which provide digital representation of the construction project \cite{karimi2021integration}. The resulting path (which is not necessarily the shortest path) can be altered with the weights of several criteria such as robot and workers safety, BIM new information requirement and sensors sensitivity to environmental features. In this step, the building semantics play an essential role on defining the start, the end and the transitional coordinates with which the robotic system plans the path. Furthermore, all along the mission, the local paths are computed based on the relevant complementary information for the low-level navigation extracted from IFC. This is essential to cope with limitations of the robot. For instance, a path planer should avoid trajectory near glass walls: they are hard to detect by many sensors. Luckily, information about wall materials can be retrieved from BIM. Among the conventional methods on path planning \cite{crespo2020semantic}, we use topological map representation in order to store the building semantics in nodes and graphs. The current paper contributions are as follows:
\begin{itemize}
    \item An optimal high-level path planner integrated with the low-level navigation (cascade navigation stack);
    \item Semantic teleoperation and navigation for autonomous UGV during the construction process;
    \item Practical implementation of the proposed system deployed on an autonomous mobile robot navigating a construction site.
\end{itemize}

The next section will summarize the inspirational works to our approach. Then, section~\ref{sec:topo} describes the generation of topological maps (hypergraphs) from IFC information. Section~\ref{sec:path} explains the IFC-based path planning algorithm in details. The setup of our field experiment to validate the proposed system is described in section~\ref{sec:field}. The results of our experimental validation are discussed in section~\ref{sec:results}. Finally, section~\ref{sec:conclusion} summarize the contributions and the next steps of our work.
\section{Related Work}
Conventional methods of indoor path planning often refer to optimal path as the shortest path calculated by various algorithms such as A* and Dijkstra's \cite{palacz2019indoor}. To enhance the performance of these planers, many studies suggested ways to leverage BIM/IFC for indoor path planning. Wang et al. \cite{wang2020bim} develop a framework for converting the BIM digital environment to a cell-based infrastructure to support indoor path planning. In this work, they emphasize on the \textit{"BIM voxelization"} process rather than the path planning problem. In another study, a BIM-based path planning strategy is used for equipment travel on construction sites \cite{song2019construction}. The authors extract the start and end points from BIM and then generate the shortest sequence of rooms for the operator, but does not support robot path planning. Ibrahim et al. \cite{ibrahim2017interactive} propose a path planning strategy based on BIM for an Unmanned Aerial Vehicle (UAV) on construction sites which uses a camera for data capturing. They use BIM geometries to define a path for outdoor environments but do not address indoor semantic robot path planning. In this direction, Follini et al. \cite{follini2020combining} utilize BIM geometries for path planning of an UGV supporting construction logistics application. Their proposed system uses a human-assisted approach in a controlled environment and is yet to thoroughly leverage BIM/IFC semantics in a construction site. In \cite{ibrahim20194d}, the optimal route for a data collection mission using an UAV is proposed. They utilize 4D BIM to identify which building spaces are expected to change during the construction phase (implemented in a simulated environment) so that the flight path navigate through those areas and collect data.

Delbrügger et al. \cite{delbrugger2017navigation} developed a framework supporting humans and autonomous robots navigation which mostly uses building geometries in a simulated environment. In \cite{nahangi2018automated}, the indoor localization of an UAV is assessed using AptilTags with their known location in a BIM-generated map. They present this work as a proof-of-concept for the use of AprilTags in indoor environment. However, due to inaccuracy of localization in their work, they improve their previous work by using Extended Kalman Filter (EKF) in their localization framework \cite{kayhani2019improved}. Another study examined the use of BIM in robot localization in which the proposed system uses a hierarchical reasoning for path planning \cite{siemikatkowska2013bim}. BIM was also demonstrated to be powerful for the identification of different paths from which a hierarchical refinement process can find the shortest path \cite{hamieh2020bim}. That work provides only high-level path (rooms sequence) with respect to BIM geometries and the integration with ROS is not studied. An approach using hypergraphs generated from IFC files was also developed in which a modified A* algorithm is able to detect the optimal path among nodes in the graph \cite{palacz2019indoor}.

In these inspiring works three aspects of the BIM potential for indoor robot path planning are yet to be thoroughly studied: (1) considering the full potential of the BIM/IFC semantic rather than only the geometry (2) integrating the high-level (rooms sequence) with the low-level sensor-based information in a full navigation stack (3) the field validation of strategy using BIM/IFC for both global and local path planning. In this paper, we cover these gaps by integrating Building Information Robotic System (BIRS) into a navigation system in ROS in order to determine the optimal path and then navigate autonomously.


\section{topological building maps created from BIM/IFC}
\label{sec:topo}
IFC data schema provides construction stakeholders with semantic information of buildings containing attributes and relationships between different entities \cite{ismail2018building}. This information can be extracted in graph database \cite{strug2017reasoning}. However, the use of that information for reasoning is complex since the IFC files encompass large amounts of data. In order to cope with this, we first identify the required data for robot navigation on construction sites, then, we extract and store the data in an XML database. The conceptual semantic relation between BIM/IFC and robot navigation is covered in a previous paper on BIRS \cite{karimi2021ontology}. We extend the hypergraphs of Palacz et al. \cite{palacz2019indoor} with the semantic and geometric information of IFC files. All the semantic information required to the global and local planers retrieved from IFC is in the form of a topological map.

As IEEE 1873-2015 \cite{7300355} defines, nodes and edges are the components of topological maps and we fill them with the following information:
\begin{itemize}
    \item Nodes contain the rooms information namely: room's name, room's unique ID, room center, room area, walls' unique IDs, wall material, last scan date, construction activity (hazard for the robot)
    \item Edges contain the doors information namely: door's unique ID, door's location, doors opening direction 
\end{itemize}

\begin{figure}[!t]
    \centering
	\includegraphics[width=3.25in]{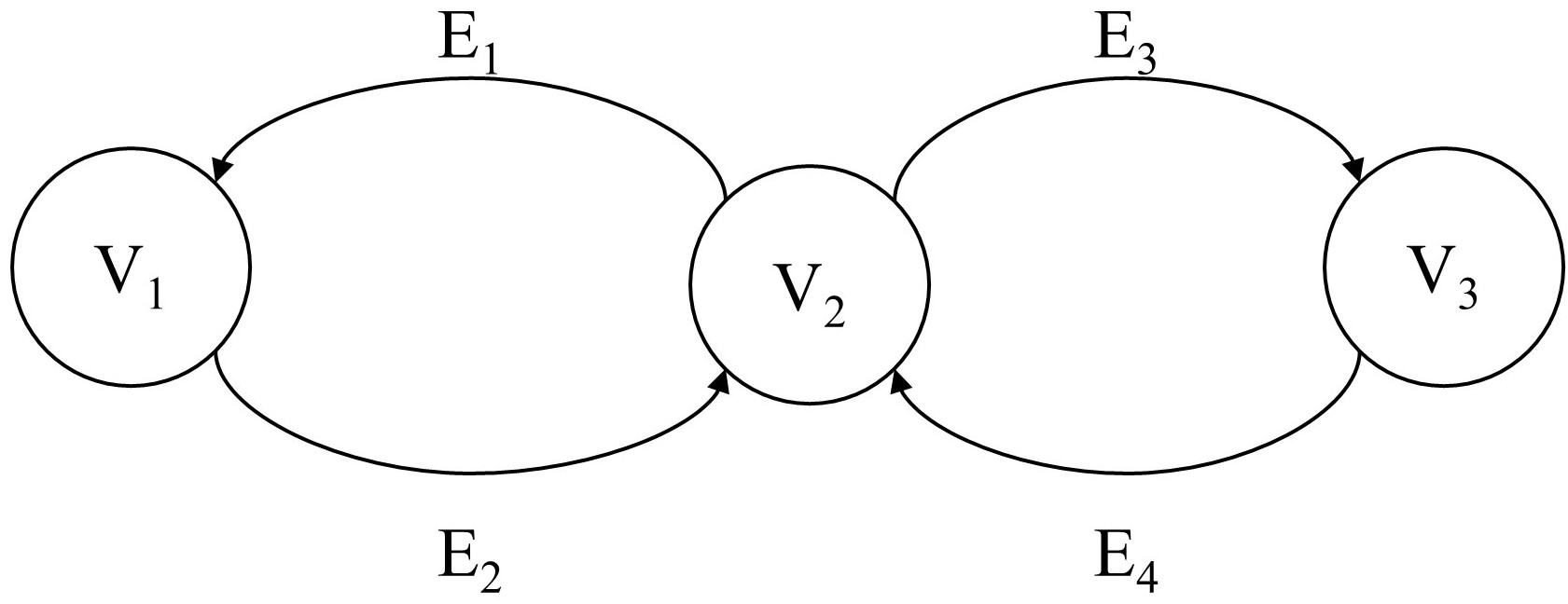}
	\caption{A directed hypergraph of $S = (V, E)$ where $V=\{V_1,V_2,...,V_n\}$ is a set of nodes and $E=\{E_1,E_2,..,E_m\}$ is a set of hyperedges. Each node $(V_i)$ is an \textit{IfcSpace} containing its relationships and each hyperedge $(E_j)$ is an \textit{IfcDoor} with its attributes extracted by BIRS \cite{karimi2021ontology}.}
	\label{fig:hypergraph}
	\vspace{-1.5em}
\end{figure}
In the hypergraph, one node is created per \textit{IfcSpace} and for each \textit{IfcSpace}, the bounding \textit{IfcWall} and \textit{IfcCurtainWall} elements are identified. With the above-mentioned information, a graph is generated as illustrated in fig. \ref{fig:hypergraph}. Then, the edges need to be attributed with the cost (weight) of passing over each (from a room to another). In this direction, $W=(W_V, W_E)$ is a pair of weights where $W_V$ and $W_E$ are the node and hyperedge weights respectively. $W_{V_i}$ is the $i$ node total weight obtained from:
\begin{equation}\label{W_V}
    W_V=w_m+w_a+w_s+w_h
\end{equation}
where $w_m$ depends on the walls material, $w_a$, on the room area, $w_s$, on the room scan-age, and $w_h$, on the room hazards. $W_{E_j}$ is the $j$ hyperedge weight obtained from:
\begin{equation}\label{W_E}
    W_E=w_d
\end{equation}
where $w_d$ depends on the door opening direction. For passing from one node to the other, there might be several paths the robot can use. The overall weight of a path (from start to end node) is as follows:
\begin{equation}\label{W}
    W = \sum_{i=1}^{n}W_{V_i} + \sum_{j=1}^{m}W_{E_j}
\end{equation}
One challenge for the robot is to be able to detect obstacles. To help the robot predict and avoid potential failures, the material properties of the walls are extracted through \textit{IfcMaterial} and its super-type \textit{IfcProduct}. The weight of each curtain wall, i.e. walls that are \emph{invisible} by design, in each node is $w_m=12$, while all others are $w_m=4$ since they can be easily detected. The time required to go through a transition node is also taken into account, i.e. bigger rooms take more time for the robot to cross. Accordingly, the weight for the rooms less than $50m^2$, between $50m^2$ to $100m^2$ and more than $100m^2$ are $w_a=2$, $w_a=8$ and $w_a=12$ respectively. Since one of the core purpose of deploying robots on construction sites is to collect data, the scanning age of all rooms is incorporated. The progress monitoring needs up-to-date data and when the robot is collecting data it can optimise its path to visit more rooms and collect more data. The scanning periods are selected according to industry needs, therefore, we assign $w_s=10$, $w_s=6$, $w_s=0$ for the scanning period of less than 1 week, between 1 week and 2 weeks, and more than 2 weeks respectively. Since the construction projects evolve constantly, the safety aspects of robot navigation are essential. In this direction, the data collection for the spaces with ongoing construction activities should be postponed to a safer moment for the robot to navigate those rooms. If the hazardous space is one of the transition nodes, an alternative route needs to be automatically planned so we assign $w_h=500$ for the weight of passing through such spaces. In this case, another path will be selected by the algorithm if there is any. If there is not an alternative safe path for the robot, the algorithms provides a warning for high-weight paths so that the supervisor of the robotic deployment is warned. The hypergraph representing building topological map enables the robotic system to find the optimal path by running an algorithm. In this paper, we use directed hypergraph (with directed hyperedges) allowing us to assign cost for door opening directions. \textit{IfcDoor} as a sub-class of \textit{IfcBuildingElement} provides the center coordinates of the doors creating hyperedges (with their coordinates) in the hypergraph. \textit{IfcDoor} also stores the opening direction through y-axis of \textit{ObjectPlacement} parameter. For pushing and pulling the door, we assign $w_d=2$ and $w_d=6$ to the hyperedge's weight respectively. This is due to difficulty for pushing and pulling the doors respectively. Ultimately, the total weight of passing one to the other is the sum of nodes weights and edges.

\section{Finding The Optimal Indoor Path}
\label{sec:path}
As Gallo et al. \cite{gallo1993directed} define, directed hypergraphs are divided into two categories according to their hyperedges namely: forward hypergraph (F-hypergraph) and backward hypergraph (B-hypergraph). The former is a directed hypergraph in which one node diverges to several nodes and the latter is the one in which several nodes converge to one node. As an example of applications, F-hypergraphs are employed for time analysis on transportation networks \cite{prakash2017finding}. Also, B-hypergraphs are used to perform deductive analysis to find the optimal path in a hypergraph. The combination of B-hypergraph and F-hypergraph is a BF-hypergraph having both divergent and convergent nodes \cite{gallo1993directed}. In topological building layouts, we deal with BF-hypergraphs since we have spaces which connect several spaces to other spaces (an example of such nodes is corridors). In addition, we intend to find the optimal path (a \textit{"deductive database analysis"} from several possible paths) based on several criteria which are represented as weights in the hypergraph, therefore, we use the \textit{"Shortest Sum B-Tree"} algorithm which finds a hypertree (subhypergraph) of the nodes as explained in \cite{gallo1993directed}. We also use \textit{additive weighting function} to calculate the cumulative weight of each possible route and then we choose the lighter route which is the optimal path for the robot.

\begin{figure}
    \centering
\noindent\begin{boxedminipage}{\linewidth}
    \begin{alltt}
Inputs:
  layout_graph : hypergraph
  tail_room, head_room : node
  door : hyperedge
  path_weight : hyperedge_total_weight
Outputs:
  semantic_path : list<node, hyperedge>
  x_y_path : list<nodes_coordinates,
  hyperedges_coordinates>
  hyperedge_total_weight : number
    \end{alltt}
\end{boxedminipage}
    \caption{Data structure for IFC-based semantic optimal path planner algorithm}
    \label{fig:data_structure}
	\vspace{-1.5em}
\end{figure}

\begin{figure*}[!h]
	\centering
	\includegraphics[trim={0 1.75cm 0 1cm}, clip,width=0.85\linewidth]{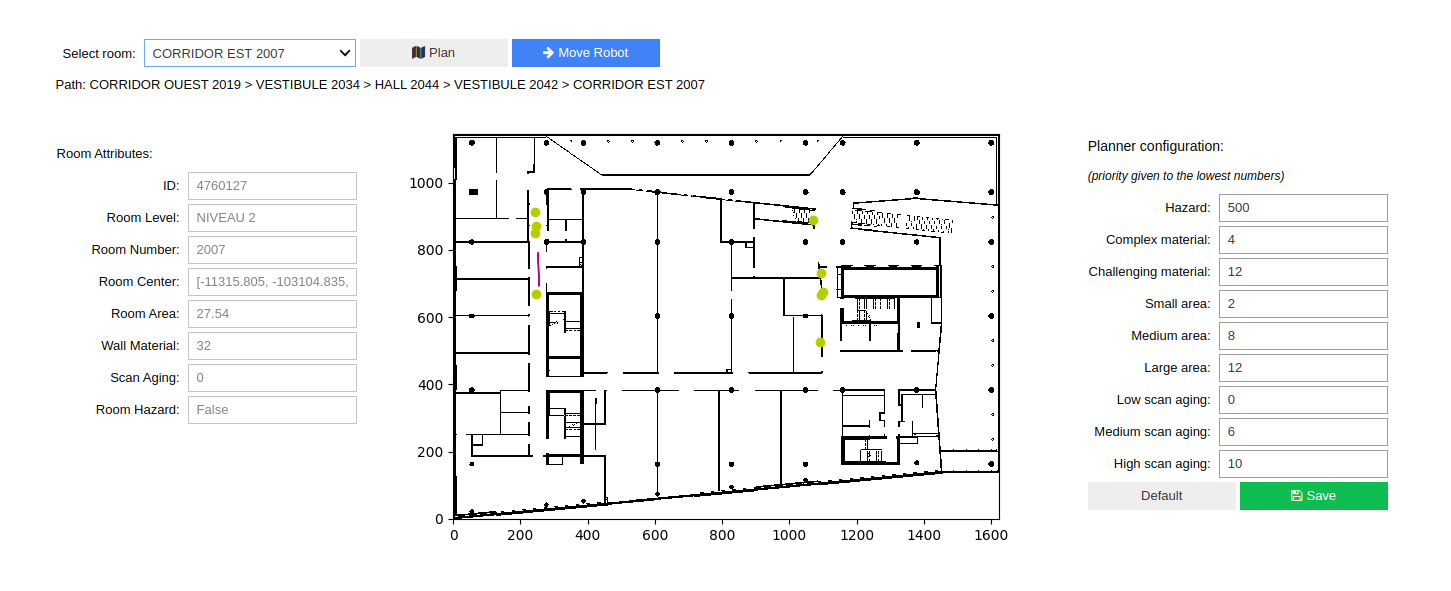}
	\caption{Semantic Graphical User Interface for the intuitive operation of the robot navigation on construction sites. The controls in the header allow selecting a destination and generating the path. The panel to the left shows the attributes of the selected room. The center contains a map of the environment, with the robot's pose in real time being represented by the purple arrow. The center points of the rooms and doors in the path are represented in the map by the yellow circles. The right panel allows the user to reconfigure the different weights that are applied to the path generation.}
	\label{fig:gui}
	\vspace{-1.0em}
\end{figure*}
In order to create the hypergraph, we first retrieve all the relevant IFC information. The process is done with a Dynamo script (a visual programming tool) to extract the IFC parameters in order to export the IFC information in a XML database. A Python script is developed to parse the XML data in order to translate meaningful data to ROS (for example, the rooms center coordinates are retrieved as strings so they need to be parsed to be integrated with the robotic system). With an hypergraph of the whole building, the user defines the start and end nodes (rooms), and let the algorithm find the optimal path. Since we are implementing BF-hypergraph, each pair of nodes is connected with two directed hyperedges together, thereby making a comprehensive B-hypergraph within the BF-hypergraph. This practice allows considering forward and backward direction in a path so that the door opening direction is considered. \textit{"Shortest Sum B-Tree"} algorithm provides the possible hyperedges from a start node to other nodes \cite{gallo1993directed}. Then, the retrieved information is used to create a sub-hypergraph from the start node to all other nodes representing all the possible paths. By giving the destination node to the sub-hypergraph, the possible paths from start to end node are identified and finally the lowest cumulative weight of the paths is retrieved. Having a set of nodes and hyperedges from the optimal path, the building information is extracted to enable semantic navigation. Each node is represented by the name of the corresponding space and the center coordinates of that room. As illustrated in fig. \ref{fig:data_structure}, the optimal path outputs a set room names, their coordinates and a set of door coordinates in the sequence of node location and hyperedge (door) location. The room names enable semantic navigation and the 2-D coordinates provides destinations one after the other.

\section{Semantic Graphical User Interface}
\label{sec:gui}

A Graphical User Interface (GUI) was developed based on BIM semantics to allow users to intuitively operate the robot and configure the path planner. The GUI connects to the ROS running in the robot and presents semantic information of the building and data from the robot in real time. The integrated high-level and low-level navigation system moves the robot to the desired destination. The GUI allows the non-expert users to work with their domain knowledge, thereby making robot deployment more intuitive and simpler. Figure \ref{fig:gui} illustrates the interface window. The GUI is developed in Python notebooks, allowing for easy integration of visualization widgets and customization.

The GUI provides the building's rooms in a drop-down list, from which the user selects a destination and then launch the path planner to find the optimal path. The center area of the GUI shows a map of the building, with the robot's pose being updated in real time, along with the paths objectives. The left panel shows the selected room's (end node) attributes. The right panel allows the user to alter the weights of each parameters of the path planner. After changing and saving the new weights, the user can generate the path again and see the results on the map. Finally, the user can click on the \emph{Move Robot} button to trigger the robot to start moving.

\section{Field deployment}
\label{sec:field}

Our approach was validated from simulation to the field with an experimental case study. The goal was to drive a mobile robot through the corridors of one of the buildings at the École de technologie supérieure, for which a complete BIM model was available, and collect data. The semantic path planner was used to generate a set of waypoints from the user inputs, then a low level A* path planner aided by a collision detection stack navigates the robot.

\begin{figure}[!t]
	\centering
	\includegraphics[width=\linewidth]{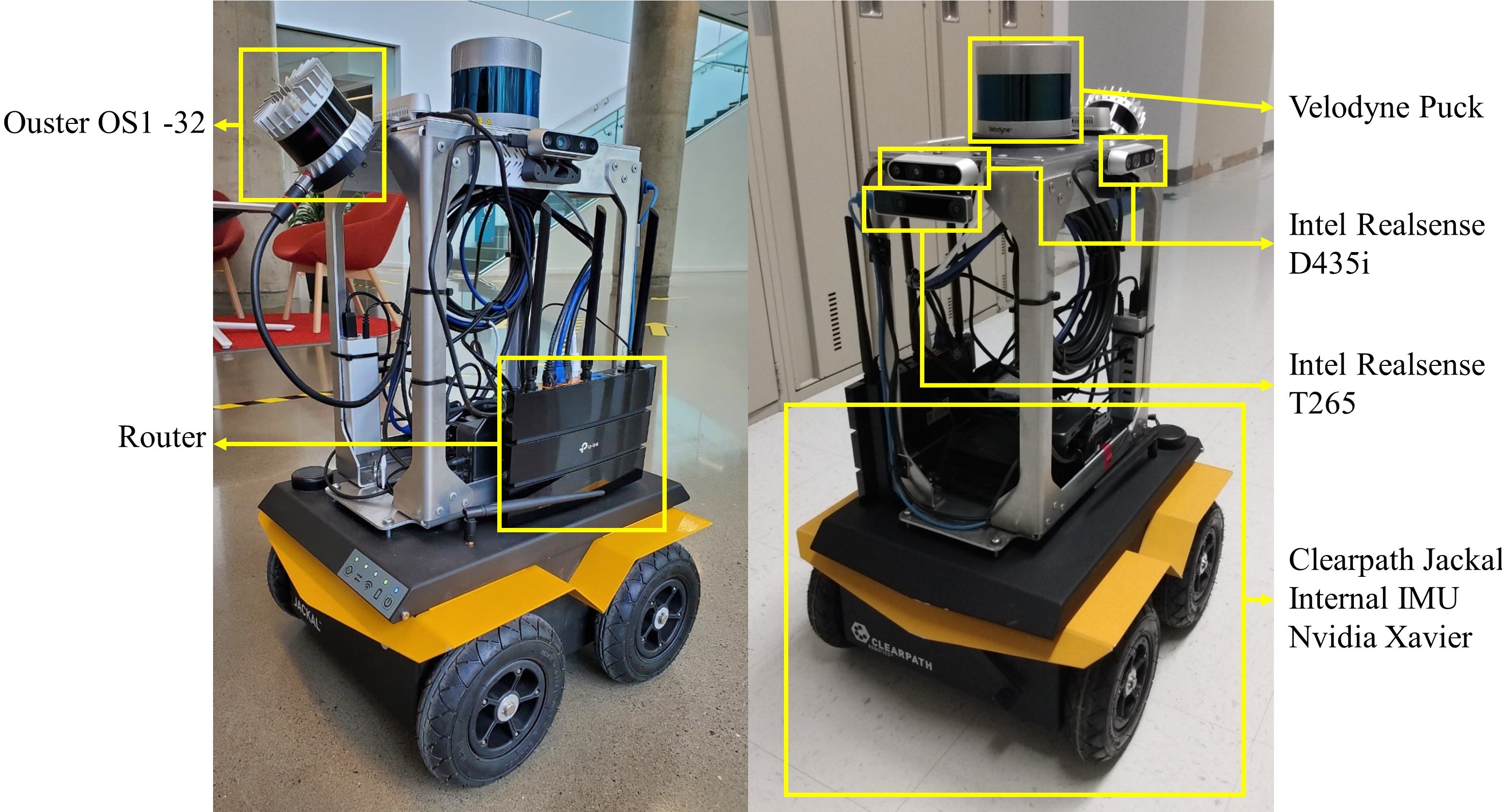}
	\caption{Mobile robot platform equipped with various sensors}
	\label{fig:robot_platform}
	\vspace{-1.5em}
\end{figure}

Our robotic platform, shown in fig. \ref{fig:robot_platform}, is built from a four-wheeled unmanned ground vehicle (Clearpath Jackal) equipped with wheel encoders, an internal IMU and an onboard NVidia Xavier computer. The Jackal is delivered with ROS nodes for control, odometry estimation (from encoders and IMU) as well as diagnostics tools provided by ROS.

The sensing system, which was envisioned for point cloud collection in construction sites, contains two LiDARs, five depth cameras and one tracking camera. The sensors are positioned in different directions to cover as much as possible of the robot's surroundings. While all sensors collect and record data of the environment, most of them are also used by the navigation stack for localization and collision avoidance. Below we present a detailed description of each sensor or group of sensors:

\begin{itemize}
    \item \textbf{Front facing cameras:} One Intel Realsense D435i depth camera and one Intel Realsense T265 tracking camera are mounted in front of the robot. The T265 software estimates the camera's pose and integrates data from the base odometry (wheel encoders and IMU), providing accurate odometry that is fed to the localization algorithm. The D435i provides depth images that are used to detect obstacles immediately in front of the robot, triggering an emergency stop;
    \item \textbf{Velodyne Puck 32MR LiDAR:} Mounted horizontally on top of the robot, it captures laser scan data from all around the robot. This information is used by the localization algorithm to estimate the robot's global position on the building map;
    \item \textbf{Depth cameras:} Three Intel Realsense D435i depth cameras are mounted pointing to the top and left and right sides of the robot. Their purpose is to collect RGB images and depth images from the walls around the robot and from the ceiling;
    \item \textbf{Ouster OS1 LiDAR:} The last sensor, an Ouster OS1 LiDAR is mounted in the back of the robot, inclined by an angle of 45 degrees in order to capture point clouds of the ceiling. Since this sensor has a large 90° field of view, it is also able to cover the walls and part of the back of the robot.
\end{itemize}

\begin{figure}[!t]
	\centering
	\includegraphics[width=3.49in]{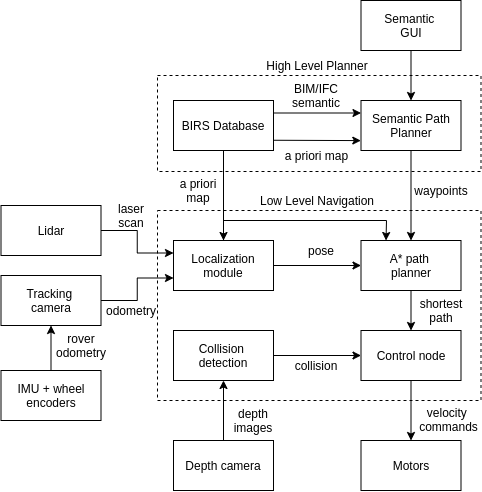}
	\caption{System Overview: A high level planner that process BIM/IFC information and user inputs is integrated to a low level navigation stack in a cascade design. The low-level module takes care of the localization, local path planning and collision avoidance tasks, while the high-level planner generates paths based on BIM/IFC semantics.}
	\label{fig:system_overview}
\end{figure}

Figure \ref{fig:system_overview} gives an overview of the system. The robot pose in the map is obtained through the use of a ROS implementation of the Adaptive Monte Carlo localization algorithm\cite{amclRos}\cite{thrun2002probabilistic}. Before deploying the robot, wall geometry information is extracted from BIM to generate an occupancy grid of the building. During the robot navigation, this map, the odometry, and the laser scan data from the horizontally mounted Velodyne LiDAR are fed to the localization algorithm, which then estimates the robot's current pose in that map. When a destination room is selected, the semantic path planner outputs the preferred path to that room as a list of waypoints, containing the center points of each room, door and corridor in the path. An A* path planner\cite{hart1968formal} then calculates the shortest path from the robot's current position to the next waypoint in the list. Velocity commands are generated from the A* path and sent to the robot's internal controller to drive it though that path.

\begin{figure}[!t]
	\centering
	\includegraphics[trim={1cm 5cm 1cm 5cm},clip,width=3.49in]{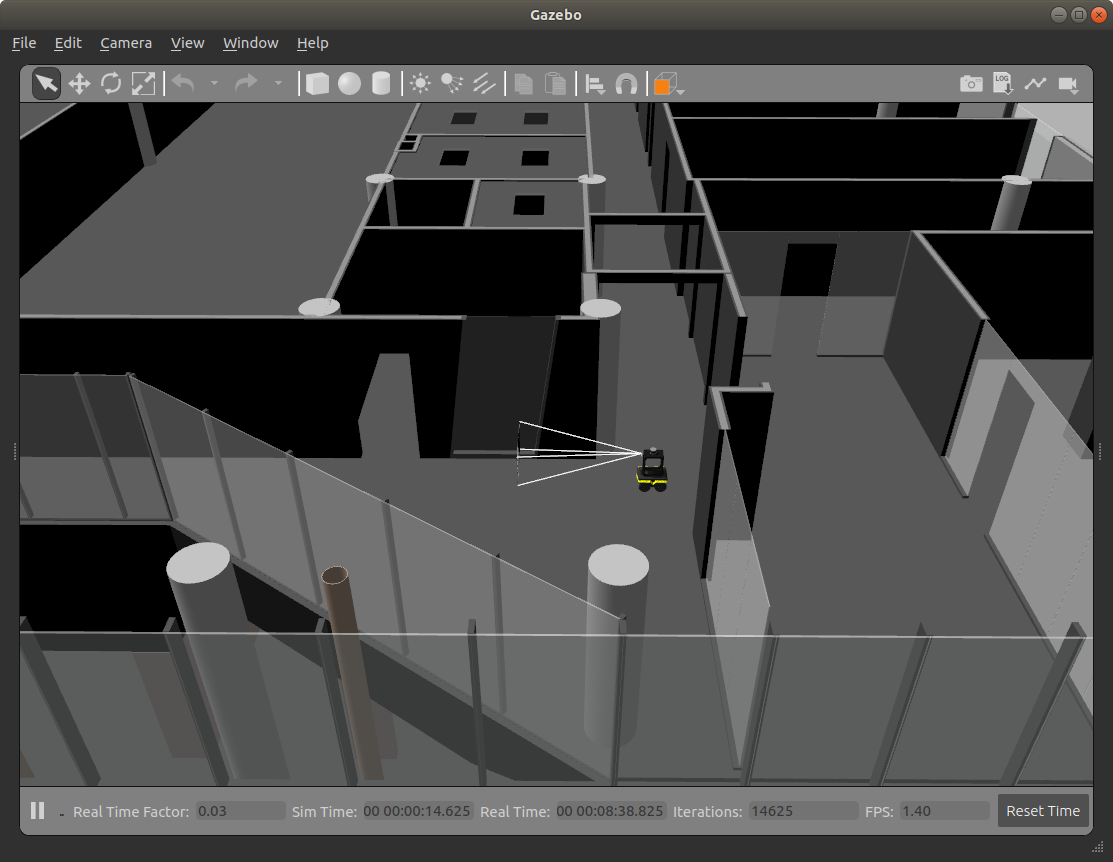}
	\caption{View of the simulated environment used to test the BIM/IFC optimal path planning approach. The building 3D model was built with geometry information extracted from the BIM. The robot model simulates the sensors and possesses the same characteristics as the real robot.}
	\label{fig:simulation}
	\vspace{-1.5em}
\end{figure}

The simulation was performed using the Gazebo Simulator. The building information is exported to create a 3D model, a digital twin. Clearpath, Gazebo and the ROS community provide all the required software packages required to generate an accurate simulation of our robotic platform. Figure \ref{fig:simulation} shows the simulated robot and its environment with different wall textures and transparency.

\section{Results}
\label{sec:results}

The experiment had two main objectives:

\begin{enumerate}
    \item Test the effectiveness of the semantic path planner in generating the optimal path to reach the destination, given the building information obtained from BIM/IFC.
    \item Test how changes in the building information affect the final path that is generated.
\end{enumerate}

\begin{figure}[!t]
\centering
\includegraphics[width=3.49in]{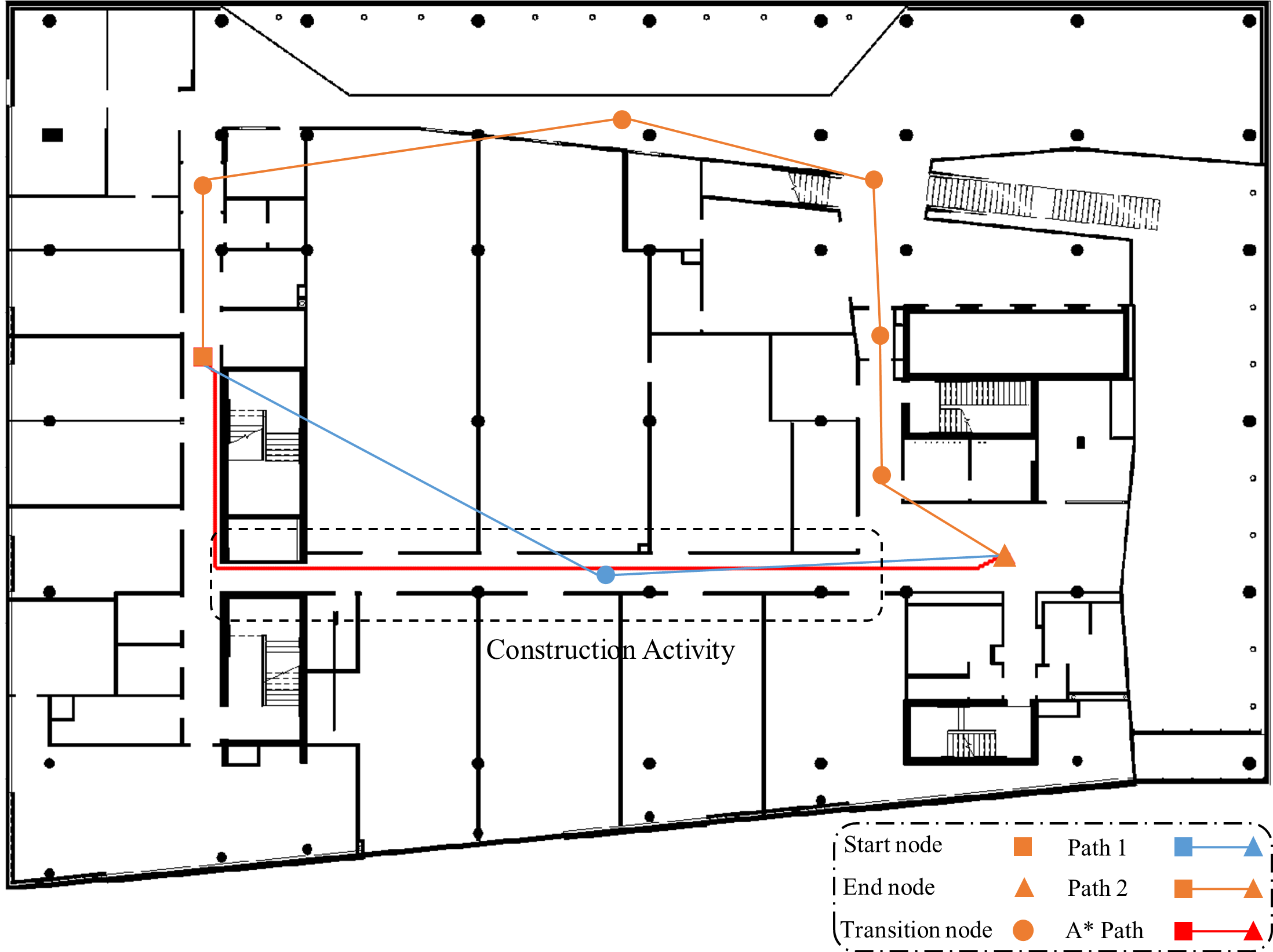}
\DeclareGraphicsExtensions.
\caption{High-level and low-level paths: A* generates the shortest path possible between start and end, not taking advantage of the BIM/IFC semantics. Path 1 has the lowest total weight among other alternatives. Path 2 is automatically generated when there is a hazard to the robot in path 1.}
\label{fig:semantic_paths}
	\vspace{-1.5em}
\end{figure}

In our case study, the robot starts in a corridor (CORRIDOR OUEST) on the west side of the building and must reach an open area (CORRIDIR EST) on the eastern part of the building. Figure \ref{fig:semantic_paths} shows the building map, and the path in red line generated by applying the A* algorithm from start to end. This is the shortest possible path between the two points, taking into consideration only the building geometry and a small safety collision radius around the robot. When the Semantic Path Planner is applied to the same scenario, a similar result is obtained as expected, represented by the blue path in fig. \ref{fig:semantic_paths}. Since there are no doors, undesirable materials or hazards in the path, the algorithm outputs a list of rooms that must be visited by the robot that represent the shortest distance from start to end. The semantic path planner provided the order of rooms' names from the start to the end as it is show in the GUI in fig. \ref{fig:gui}. Therefore, the user operating the robot can intuitively track the path from the data collected. In this direction, the as-built data can be directly compared to the as-planned since the path is recorded semantically. Also, the waypoints of rooms' center coordinates and doors' center coordinates are provided by the semantic path planner. If there is a door made of materials invisible to sensors (such as glass), the complementary door coordinates helps for safer, smarter, precise data collection. Following this, the A* algorithm finds the shortest path between the waypoints. 

In a second run, the building information was altered to include a construction operation carried out in the area highlighted with a dashed box in fig. \ref{fig:semantic_paths} (not visible in the GUI). Since the construction activity represents a hazard with a high cost for the Semantic Path Planner, a different path passing through another corridor is automatically selected, as illustrated by the orange path. Nevertheless, the high cost of the shortest path triggered a warning in the system indicating a hazard to the user through the semantic GUI. Therefore, the user can understand the risks associated with navigation through an active construction area and decide whether to scan the environment or postpone it to a safer time. The orange path was automatically generated, although it is not the shortest path, as the optimal path from the default parameters mentioned in section \ref{sec:path}. This path passes along a large curtain wall invisible to the robot's sensors. The additional semantic information provided by the BIRS is given to the robot as well as the BIM occupancy grid so it contributes to collision avoidance with the wall. The GUI provides the user with the scan aging of the rooms so the user can decide which rooms to select as the destination for data collection. This allows the users to run multiple data collection mission with the robot which increases the efficiency of robot deployment on construction sites. As illustrated in fig. \ref{fig:semantic_paths}, the integrated BIM-ROS information provides a cascade navigation system on construction sites enabling autonomous and accurate data collection of the spaces scanned.

\section{Conclusion}
\label{sec:conclusion}
This paper presented a semantic path planner that uses building information from IFC data schema to generate optimal paths for safe and efficient navigation of autonomous robots on job sites during the construction phase. We used the BIRS for extracting building information from IFC represented in a hypergraph structure. Path planning algorithms can then be used to calculate optimal paths in this graph given the building information. Weights are designated to each connection in the path to represent how different conditions can affect the robot's navigation and to prioritize paths with more desired characteristics. The optimal semantic path is then integrated with low-level navigation system and A* algorithm is used to calculate the shortest path within the optimal path. The effectiveness of the path planning to generate different paths given different conditions was shown in a simulated and real life case study.

This algorithm can be extended in the future to take into consideration Mechanical, Electrical and Plumbing (MEP) semantics for data collection. Different locations can be added based on the kind of information needed at a specific time of construction through the GUI in order to provide the robot more destinations to collect data. Therefore, the high-level path planning algorithm would provide a more efficient route for data collection as well as semantic navigation. Also, this paper provided semantic navigation of mobile robot on construction sites, therefore, a user study will be conducted in order to assess the usability of the semantic navigation approach.

\section*{Acknowledgment}
The authors are grateful to the Natural Sciences and Engineering Research Council of Canada for the financial support through its CRD program 543867-2019, to Mitacs for the support of this field study as well as Pomerleau; the industrial partner of the ÉTS Industrial Chair on the Integration of Digital Technology in Construction.

\bibliographystyle{IEEEtran}
\bibliography{IEEEabrv,bibtex/bib/main}

%





\end{document}

%% file: macros.tex
\newcommand{\be}{\begin{equation}}
\newcommand{\ee}{\end{equation}}
\newcommand{\beq}{\begin{equation}}
\newcommand{\eeq}{\end{equation}}
\newcommand{\bed}{\begin{displaymath}}
\newcommand{\eed}{\end{displaymath}}
\newcommand{\beqa}{\begin{eqnarray}}
\newcommand{\eeqa}{\end{eqnarray}}
\newcommand{\beqann}{\begin{eqnarray*}}
\newcommand{\eeqann}{\end{eqnarray*}}
\newcommand{\bseq}{\begin{subequations}}
\newcommand{\eseq}{\end{subequations}}

\newcommand{\ba}{\begin{array}}
\newcommand{\ea}{\end{array}}

\newcommand{\real}{\hbox{I \kern -.5em R}}

